\documentclass[letterpaper]{article} 
\usepackage{aaai25}  
\usepackage{times}  
\usepackage{helvet}  
\usepackage{courier}  
\usepackage[hyphens]{url}  
\usepackage{graphicx} 
\usepackage{natbib}  
\usepackage{caption} 
\usepackage{algorithm}
\usepackage{algorithmic}

\usepackage{amsmath}
\usepackage{graphicx}
\usepackage{microtype}
\usepackage{array}
\usepackage{tabularx}
\usepackage{subcaption}
\usepackage{multirow}
\usepackage{soul}

\usepackage{newfloat}
\usepackage{listings}
\DeclareCaptionStyle{ruled}{labelfont=normalfont,labelsep=colon,strut=off} 
\floatstyle{ruled}
\newfloat{listing}{tb}{lst}{}
\floatname{listing}{Listing}
\nocopyright 

\title{PMMT: Preference Alignment in \\ Multilingual Machine Translation via LLM Distillation}
\author{
    Shuqiao Sun\textsuperscript{\rm 1}\equalcontrib,
    Yutong Yao\textsuperscript{\rm 1}\equalcontrib,
    Peiwen Wu \textsuperscript{1},
    Feijun Jiang \textsuperscript{1},
    Kaifu Zhang \textsuperscript{1}
}
\affiliations{
    \textsuperscript{\rm 1}Alibaba International Digital Commerce Group, AI Business

}

\usepackage{bibentry}

\begin{document}

\maketitle

\begin{abstract}
    Translation is important for cross-language communication, and many efforts have been made to improve its accuracy. 
    However, less investment is conducted in aligning translations with human preferences, such as translation tones or styles.
    In this paper, a new method is proposed to effectively generate large-scale multilingual parallel corpora with specific translation preferences using Large Language Models (LLMs).
    Meanwhile, an automatic pipeline is designed to distill human preferences into smaller Machine Translation (MT) models for efficiently and economically supporting large-scale calls in online services.
    Experiments indicate that the proposed method takes the lead in translation tasks with aligned human preferences by a large margin. Meanwhile, on popular public benchmarks like WMT and Flores, on which our models were not trained, the proposed method also shows a competitive performance compared to SOTA works.
\end{abstract}

%

\section{Introduction}

Recently, translation methods are improved with technological advancements. However, most of them pursue the accuracy criteria and neglect the vivid and nuanced expressions of languages \citep{DBLP:journals/corr/abs-2207-04672,DBLP:journals/jmlr/FanBSMEGBCWCGBL21}. 

By analyzing the dialogues from our online services, we noticed that "human preferences" in expression vary drastically among users. 
For example, some of them want to translate "Can you" to "¿Podría" instead of "¿Puede" to provide a more polite tone. This is hard to achieve using traditional quality estimation (QE) filters or the intervention techniques because the Spanish word varies in different condition and the translation preference is conflict to the semantic mapping.

Although many researches are conducted to address the problem of preference alignment \citep{DBLP:journals/corr/abs-2401-12873, DBLP:journals/corr/abs-2402-11525, DBLP:conf/emnlp/WuTQLL18}, challenges remain in aligning these specific customized preferences to translation besides aligning to common rules like the 3H criteria (helpfulness, harmless, and hallucination).

Firstly, despite the abundance of datasets for general translation \citep{DBLP:journals/corr/abs-2305-04118,DBLP:conf/emnlp/MichelN18}, there is a paucity of customized corpora \citep{DBLP:journals/corr/abs-2309-11674,DBLP:journals/corr/abs-2401-08417}. 
Possible reasons are: 1) preferences vary significantly across applications and are hard to gather, and 2) constructing large-scale parallel corpora is costly for diverse preferences.

Secondly, due to cost, latency, and stability concerns, using small models for online translation services is more practical. However, the translation from a pre-trained small model of a given source text can hardly be changed by prompts.
As a result, when a new preference is required, update the models can be costly due to training corpus re-generating and manual model re-training.

To address these challenges, we propose an automatic data generation method that can be easily generalized to large-scale parallel corpus production with specific preferences.
Firstly, a small seed dataset is built through multiple training resources, on which a translation LLM is trained to generate more candidate translations from the new source texts.
Meanwhile, a reward model (RM) is trained on another small dataset which is aligned to human preferences. 
Finally, the candidate translations are filtered by the Best-of-N (BoN) strategy \citep{DBLP:journals/corr/abs-2112-09332}.
With the growth of source text scale, ceaseless high-quality parallel data will be produced through the pipeline automatically.

To relieve the cost of alignment to fickle preferences, we propose an approach to automatically distill knowledge from LLMs to smaller models.
In this pipeline, only few examples that represent new preferences are required to update the seed preference data and all small translation models will be updated automatically.

\begin{figure*}
    \includegraphics[width=\linewidth]{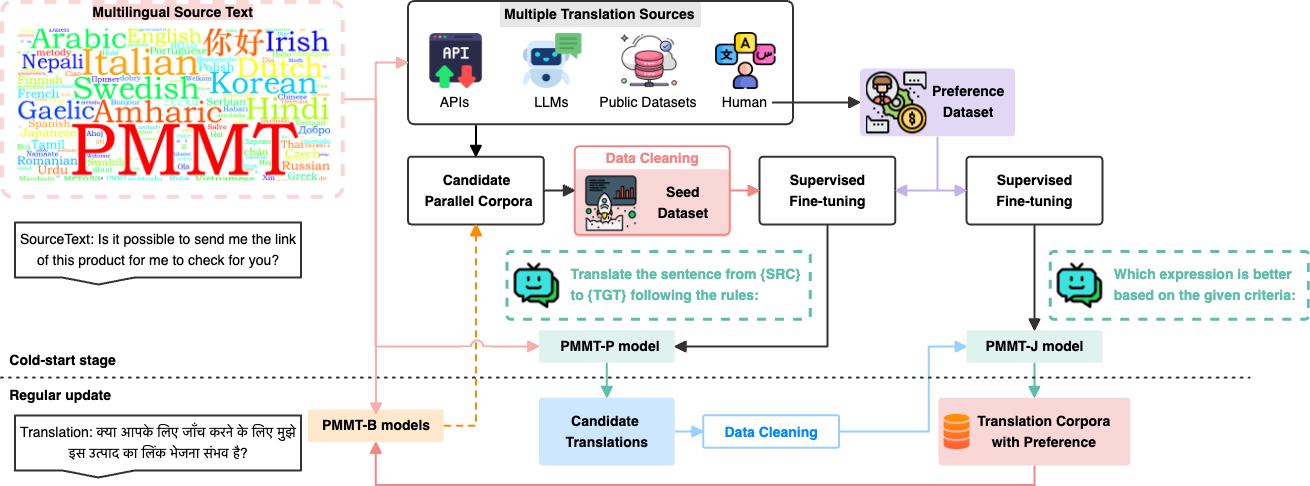}
    \caption {Pipeline of the proposed PMMT method.}
    \label{fig:method}
  \end{figure*}

Evaluation are conducted on customized dataset and public benchmarks including WMT23 \citep{DBLP:conf/wmt/KocmiABBDFFFGGH23} and flores \citep{DBLP:journals/corr/abs-2207-04672}.
Experimental results show that the proposed method can generate first-class translations in general tasks while takes the lead by a large-margin in tasks with specific human preferences. 
Figure.~\ref{fig:method} illustrates an overview of the proposed method and details are introduced in the following sections.

In General, the main contributions of this paper are:
\begin{itemize}
    \item Proposing a method to generate high-quality, large-scale parallel translation corpora that are aligned to customized human preferences.
    \item Developing an approach to distill knowledge from LLMs to smaller models through automatic model update.
\end{itemize}

\section{Related Work}
\label{sec:related_works}

\subsection{Translation Models}

Multilingual Machine Translation (MMT) has been extensively researched and numerous methods are proposed.
Among them, M2M \citep{DBLP:journals/jmlr/FanBSMEGBCWCGBL21} is one of the most representative works, which achieves translation between two arbitrary languages without using English as a bridge. 
Later, NLLB \citep{DBLP:journals/corr/abs-2207-04672} was proposed to expand the translation capabilities on low-resource languages.

Recently, LLMs are widely leveraged by MMT methods.
Some of them target in achieving MMT through prompt engineering \citep{DBLP:journals/corr/abs-2305-04118} while others prefer to improve the translation accuracy through model training.
For example, the English-based model LLaMA-13B \citep{DBLP:journals/corr/abs-2302-13971} is augmented on Chinese corpora for a bilingual-central translation \citep{DBLP:journals/corr/abs-2305-18098}.
However, multilingual parallel corpus can be scarce sometimes, so a two-stage training method \citep{DBLP:journals/corr/abs-2309-11674} is proposed, where the model is firstly trained on large-scale monolingual corpora to learn the expression in single languages, and then use small-scale human-annotated parallel corpora to learn the mapping between languages.
Although these models have demonstrated powerful capabilities in MMT tasks, aligning them with human preferences is expensive due to dataset generation and training cost. 

\subsection{Dataset Generation}

The quality and quantity of the parallel corpus determine the accuracy upper bound of a translation model \citep{DBLP:journals/corr/abs-2404-00929} and many efforts have been made to generate large-scale high-quality datasets. 

For example, He et. al attempt to utilize the multilingual ability of LLMs with multi-step prompt engineering approaches for data generation \citep{DBLP:journals/corr/abs-2305-04118}. MTNT \citep{DBLP:conf/emnlp/MichelN18} produces manual-proofreading high-quality parallel corpora through pre-filtering and noise identification. SentAlign \citep{DBLP:conf/emnlp/SteingrimssonLW23} utilizes a variant of the Dijkstra algorithm and the LaBSE model semantically align the source and target texts, thereby improving the accuracy and scalability when processing large-scale data.
Works are also done to extract high-quality parallel corpora from noisy translation datasets following a few-shot prompt with Chain-of-Thought (CoT) reasoning \citep{DBLP:conf/emnlp/BoldingLDLM23}. Many QE method are also proposed for data cleaning \citep{DBLP:journals/corr/abs-2402-17733,DBLP:journals/corr/abs-2309-11674,DBLP:conf/wmt/BlainZRGKSSVJAO23}.

Although these methods can provide high-performant results, some challenges remain. 
First of all, most of them only target on general linguistic rules and is clumsy for aligning with capricious customized human preferences. Meanwhile, considering the large quantity of the raw data, directly use large QE models for data cleaning is not practical in most cases.
  
\subsection{Preference Alignment}

Reinforcement Learning (RL) has been proved to be effective in numerous preference alignment applications, including machine translation \citep{DBLP:conf/emnlp/WuTQLL18}.
Methods are proposed to elevate the BLEU score \citep{DBLP:conf/wmt/Post18} of translation models with strategies such as leveraging a distributed policy gradient algorithm \citep{DBLP:journals/corr/abs-2207-08583}.

Additionally, several works have been conducted based on Reinforcement Learning with Human Feedback (RLHF) \citep{DBLP:conf/nips/Ouyang0JAWMZASR22} to achieve preference alignment according to higher human standards.
An MBR decoder with DPO \citep{DBLP:conf/nips/RafailovSMMEF23} is proposed \citep{DBLP:journals/corr/abs-2311-08380} to refine translations based on human feedback, and a CPO method for reference-free translation fine-tuning \citep{DBLP:journals/corr/abs-2401-08417}.
Additionally, more subsequent methods \citep{DBLP:journals/corr/abs-2401-12873, DBLP:journals/corr/abs-2402-11525} have been proposed to align higher-level human preferences.

However, preferences defined in the above-mentioned methods still mainly refer to the translation accuracy under general applications, while neglecting higher-order human preferences such as faithfulness, expressiveness, and elegance (referring to the satisfaction of customized needs).

\section{Approach}
\label{sec:approach}

The proposed method mainly consists of two parts: the cold-start stage and the regular update stage.
  
During the first stage, a small-scale seed dataset is built by leveraging multiple translation resources.
The translations are cleaned before training the LLM (named as PMMT-P, where "P" stands for "Producer"), which is used to generate the candidate translations for following steps.
At the same time, an RM (named as PMMT-J, where "J" stands for "Judge") is trained on a small customized preference dataset, and is thereafter used to select translations that match specific preferences.

During the second stage, numerous parallel data are generated by PMMT-P and PMMT-J with the update of the source texts, which contain both the translation norms learned by PMMT-P and the preferences learned by PMMT-J.
Then, smaller models (named as PMMT-B) are automatically trained on the generated corpora and thus distill the knowledge from LLMs into the translation.

\subsection{Seed Dataset}
\label{sec:data_clean}

The seed dataset is used to train the PMMT-P and the source texts are from our online e-commerce conversations (see Figure.~\ref{fig:method} for reference). Multiple translation resources are utilized to maintain the diversity of outputs from PMMT-P: translation APIs (Google translation and Damo translation), sophisticated AI models \citep{DBLP:conf/nips/BrownMRSKDNSSAA20, DBLP:journals/corr/abs-2303-08774}, and human annotators. 

To elevate the accuracy of the seed dataset, a series of data cleaning strategies are designed and applied to multiple stages of PMMT.
The main strategies are introduced below and they are applied in a descending order with respect to computational complexity to reduce the total cost.

\paragraph{Length.}
Empirically, the length of the translation usually approximately matches that of the source text.
Therefore, a length difference $ f_{\ell} $ is calculated to filter the translations which are much longer or shorter than the source text:

\begin{equation}
    \label{eq:length}
    f_{\ell}(S, T, \alpha) = \max\left(\frac{\ell(S)}{\ell(T)}, \frac{\ell(T)}{\ell(S)}\right) \leq \alpha
  \end{equation}
  
\noindent where $\ell(\cdot)$ calculates the length of the source text $S$ and the target text $T$, and $\alpha$ is the threshold determined according to the language cluster: $\alpha=2$ for texts within a same language cluster (e.g., English and Spanish), and $\alpha=3$ for texts from different clusters (e.g., English and Chinese).

\paragraph{Language.}
Translation resources might produce incorrect translations in the form of wrong languages.
We use two python language detection tools to check the translation languages: lingua and langid. A translation is considered correct when it passes any one of the tools.

\paragraph{Danger words.}
Content hallucinations from translation resources such as LLMs can be filtered by certain keywords such as "->", "to {target language}", "translat", "my sentence is", etc.
When a "danger word" is detected in the translation, the whole translation (along with source text) will be removed from the corpus.
For non-English target languages, the outputs will first be translated into English using a relatively reliable method for word checking. 
Additionally, to avoid destroying icons and emojis during the process, emojis will also be verified using an emoji list.

\paragraph{Number.}
The presence of inaccurate numbers in the training corpus could significantly affect the performance of 
the translation models trained on this dataset, which can be fatal to applications like e-commerce, where order numbers are pivotal to providing satisfactory services.
As a result, we specifically checked for number consistency between $S$ and $T$ using regex match rules.
If the numbers equal to each other, then the translation $T$ is regarded to pass the number check.

\paragraph{Similarity.} Semantic consistency between $S$ and $T$ is checked by a similarity score $f_{\text{sim}}$ defined by Equation.~\ref{eq:sim}, where $\mathcal{T}$ is the translation process:

\begin{equation}
    \label{eq:sim}
    f_{\text{sim}}(S, \mathcal{T}(T)) = \cos(\mathcal{M}(S), \mathcal{M}(\mathcal{T}(T)))
  \end{equation}
  
\noindent where $\mathcal{M}$ is the Minilmv2 model \citep{DBLP:conf/acl/WangBHDW21} used for embedding computation, $\cos$ is the cosine similarity function, and $\mathcal{T}$ represents the translation process.

\begin{figure*}
    \includegraphics[width=\linewidth]{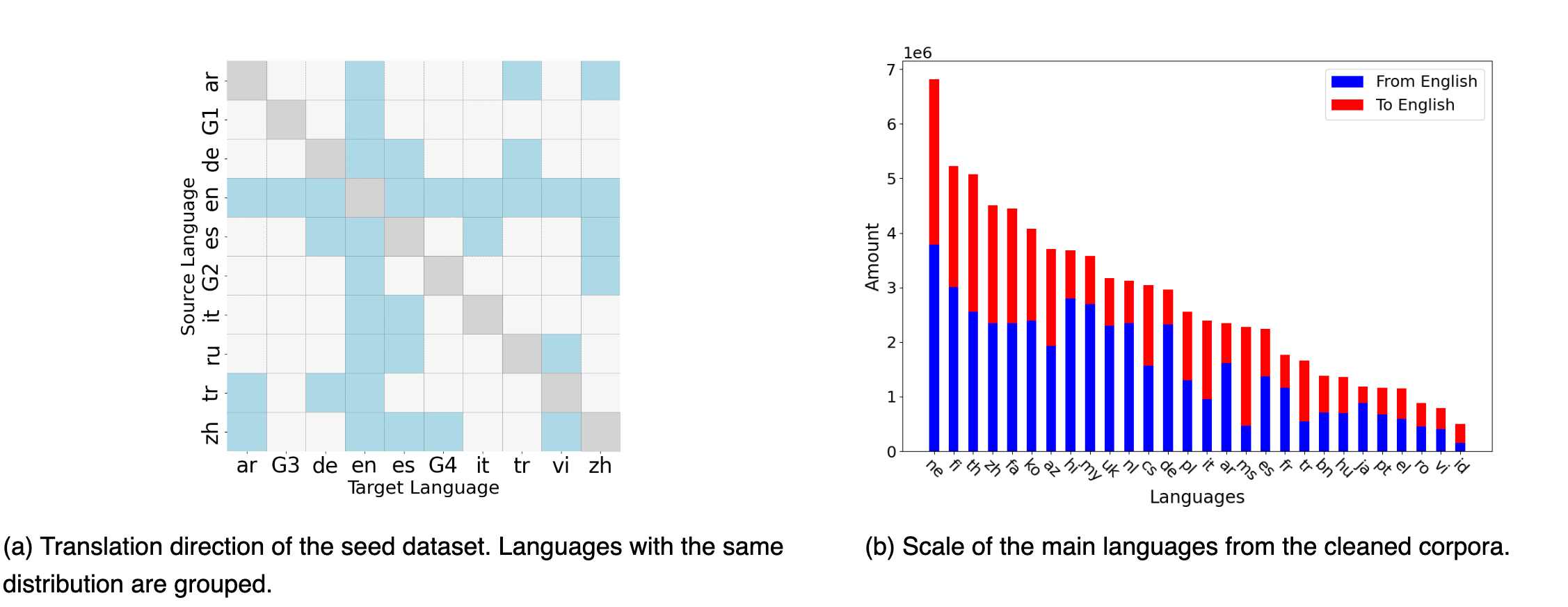}
    \caption{Dataset distribution. Gray blocks: no need for translation. Blue blocks: have data; White blocks: no data. G1/G3: az, bn, cs, el, fa, fi, hi, hu, id, ko, ms, my, ne, nl, pt, ro, th, uk. G2/G4: fr, ja, pl, vi.}
    \label{fig:data_dist}
\end{figure*}

The final seed dataset contains over 28 languages, and Figure.~\ref{fig:data_dist} (a) illustrates an overview of the language distribution.

\subsection{PMMT-P Model}
The PMMT-P model is an LLM trained on the seed dataset. Its mission is to produce raw candidate multilingual translations in multiple tones and styles. Those translations that matches the customized human preference will be chosen by the judge model PMMT-J in the following pipeline.
The PMMT-P model in this paper is fine-tunned from the pre-trained LLaMA2-13B \citep{DBLP:journals/corr/abs-2307-09288} model using the loss function $\mathcal{L}(\theta)$:

\begin{equation}
    \mathcal{L}(\theta) = - \sum_{t=1}^{T} \log p(y_t | y_{<t}, x; \theta)
    \label{eq:loss_sft}
\end{equation}
\noindent where $\theta$ represents the model parameters, and $p(y_t | y_{<t}, x; \theta)$ denotes the probability of the correct token $y_t$, given the preceding tokens $y_{<t}$ and the input $x$.

\subsection{PMMT-J Model}
\label{sec:approach_pmmt_j}

The PMMT-J model is trained on a small-scale seed preference dataset.
In this paper, we use human preferences for colloquial conversations with politeness for online e-commerce conversations as an example, but the proposed method can be easily expanded to accommodate other customized preferences.

The seed preference dataset is built from two sources: the seed dataset and human translators.
For data from the seed dataset, GPT-4 \citep{DBLP:journals/corr/abs-2303-08774} is used to label the chosen and rejected answers following these rules:

\paragraph{Correctness.} The translation should be correct in grammar and orthography.
In instances where the source text contains minor typos, the translation is still expected to convey the correct meanings.

\paragraph{Politeness.} The translation should be in a polite tone, taking into account different cultures. For example, a seller might address the customer as "\raisebox{-0.2\height}{\resizebox{!}{\ht\strutbox}{\includegraphics{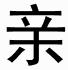}}}" or "\raisebox{-0.2\height}{\resizebox{!}{\ht\strutbox}{\includegraphics{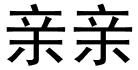}}}" in Chinese, which should be translated to "dear" or "dear customer" in English, rather than the literal "kiss" or "kiss kiss" (which represents the apparent meaning of the source text).

Furthermore, some countries may adhere to stricter guidelines regarding formalities and expect the seller to address the customer in a more respectful manner rather than using an overly friendly tone. For example, a Japanese customer would prefer to be addressed as "\raisebox{-0.2\height}{\resizebox{!}{\ht\strutbox}{\includegraphics{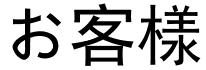}}}" instead of "\raisebox{-0.2\height}{\resizebox{!}{\ht\strutbox}{\includegraphics{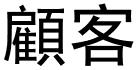}}}" since it is more polite and courteous.

\paragraph{Special cases.} There can be some customized translation requirements. 
For example, "may I" and "can I" in English should be translated into "¿Puedo" or "¿Puedes" in Spanish generally. 
However, as is mentioned above, some users would prefer to rewrite them into "¿Podría" to convey a more polite tone.

All these preferences are considered while building the seed preference dataset. Additional preferences can also be added by introducing a small amount of new parallel data which can reflect the new preferences.
The seed preference dataset and the PMMT-J model are updated only when the preference varies to maintain a more coherent filtering rule and save cost.

\subsection{PMMT-B Model}
The PMMT-B model is trained on cleaned large-scale parallel multilingual corpora with human preferences, which are generated by PMMT-P model and filtered by the PMMT-J model.
The cleaned multilingual corpora contain about 1 billion distinct entries, and the main constitution is shown in Figure.~\ref{fig:data_dist} (b).
The model structure of PMMT-B is the same as that described in \citep{DBLP:journals/jmlr/FanBSMEGBCWCGBL21}, which is a standard encoder-decoder model.

\section{Experiments}

\subsection{Training}
\label{sec:training}

\subsubsection{Settings}

All models are trained on A100 GPUs using PyTorch \citep{DBLP:conf/nips/PaszkeGMLBCKLGA19}.
The PMMT-B models are trained with the conventional training approach for standard transformer models \citep{DBLP:conf/emnlp/WolfDSCDMCRLFDS20} while PMMT-P and PMMT-J are trained within an LLM training framework \citep{DBLP:journals/corr/abs-2403-13372}.

\subsubsection{PMMT-P Model}

The PMMT-P model is trained on the seed dataset of all language pairs Simultaneously, where one language pair refers to translations from a specific language (X) to English and vice versa.
The model is only updated when new languages are added or reaches a preset update time (which is 1 month in our practice).
The model is trained for 5 epochs on the seed dataset at a start learning rate of $10^{-5}$.

\subsubsection{PMMT-J Model}
The PMMT-J model is trained on the seed preference dataset, where each source text is paired with multiple translation outputs that reflect different preferences. 
During the model training, the value head of PMMT-J scores each output generated from the base LLM.
The objective of the training process is to maximize the gaps between chosen and rejected translations
\citep{DBLP:conf/nips/Ouyang0JAWMZASR22}:

\begin{equation}
    \begin{aligned}
      L_{\text{JM}}(\theta) = & -\mathbf{E} \left[
      \log \left( \sigma\left(r_\theta (S, T^+) \right.\right.\right.\\
      & \left.\left.\left. - r_\theta (S, T^-)\right) \right) \right]
    \end{aligned}
    \label{eq:rm_loss}
  \end{equation}
\noindent where $\sigma$ stands for the sigmoid function, $r_\theta (S, T)$ is the scalar output of PMMT-J with parameters $\theta$.
$T^+$ denotes the chosen translations, and $T^-$ stands for the rejected translations.

\begin{table*}
    \centering
    \begin{tabular}{r|c|c|c|c|c|c}
    \hline
    \multirow{2}{*}{Models} & \multicolumn{2}{c|}{Customized Data} & \multicolumn{2}{c|}{Flores200 } & \multicolumn{2}{c}{WMT23} \\ 
    \cline{2-7}
     & BLEURT & COMET & BLEURT & COMET & BLEURT & COMET \\ 
    \hline
    M2M-418M \citep{fan2020englishcentric} & 0.69 & 0.81 & 0.66 & \textbf{0.83} & 0.64 & 0.75  \\
    NLLB-600M \citep{nllbteam2022language} & 0.72 & 0.83 & \textbf{0.71} & 0.76 & 0.66 & \underline{\emph{0.77}} \\
    LLaMA2-13B \citep{touvron2023llama2} & 0.67 & 0.76 & 0.64 & 0.74 & 0.65 & 0.75  \\
    Qwen1.5-14B \citep{bai2023qwen} & 0.72 & 0.79 & 0.67 & \underline{\emph{0.80}} & \underline{\emph{0.67}} & 0.76  \\
    ALMA-R-13B \citep{DBLP:journals/corr/abs-2401-08417} & 0.66 & 0.78 & 0.57 & 0.64 & 0.58 & 0.68  \\
    TowerInstruct-13B \citep{DBLP:journals/corr/abs-2402-17733} & 0.59 & 0.72 & 0.58 & \textbf{0.83} & 0.50 & 0.67  \\
    \hline
    PMMT-B-418M & \underline{\emph{0.83}} & \textbf{0.90} & 0.62 & 0.75 & 0.62 & 0.74 \\
    PMMT-P-13B & \textbf{0.87} & \underline{\emph{0.89}} & \underline{\emph{0.70}} & 0.76 & \textbf{0.68} & \textbf{0.78}  \\
    \hline
    \end{tabular}
    \caption{Translation performance evaluation results. The best and second-best results are highlighted in \textbf{bold} and \underline{\emph{italic}}, respectively.}
    \label{tab:trans_acc}
\end{table*}
  
\subsubsection{PMMT-B Model}
To ensure translation stability and the update flexibility of the online serving, all PMMT-B models are trained on individual language pairs, meaning that each PMMT-B model can only translate between English and another language.
Language pairs that are not covered by direct translation PMMT-B models will be translated using English as the bridge.

The models are trained for 15-20 epochs on the generated corpus, and the number of epochs is determined by the dataset scale empirically. 
The loss function $ \mathcal{L}(\theta) $ used to train the PMMT-B models is defined as follows \citep{DBLP:journals/jmlr/FanBSMEGBCWCGBL21}:

\begin{equation}
  \mathcal{L}(\theta) = -\frac{1}{N} \sum_{i=1}^{N} \sum_{t=1}^{T_i} \log P(y_{i,t} | y_{i,<t}, x_i; \theta)
\end{equation}
\noindent where $\theta$ denotes the model parameters, $N$ is the number of training samples, $ T_i $ indicates the length of the $i$-th target sequence, $ y_{i,t} $ stands for the $t$-th token in the $i$-th target sequence, $ y_{i,<t} $ refers to the sequence of tokens in the $i$-th target sequence preceding the $t$-th token, $ x_i $ is the $i$-th source sequence, and $ P(y_{i,t} | y_{i,<t}, x_i; \theta) $ denotes the predicted probability of the token $ y_{i,t} $ given the preceding tokens $ y_{i,<t} $ and the source sequence $ x_i $, parameterized by $ \theta $.

The PMMT-B models are regularly updated at a preset frequency using new parallel corpora generated by PMMT-P and PMMT-J from added source texts.
This practice ensures that the deployed PMMT-B models are aware of new words and preferences.

\subsection{Evaluation}
\subsubsection{Datasets}
As is mentioned above, our method is proposed to provide accurate translations with human preferences in expressions like tones, styles, etc.
Therefore, evaluation of the proposed method is conducted in two fields: 
\begin{itemize}
    \item Public benchmarks: To evaluate the models' basic translation ability like accuracy and fluency.
    \item Customized test set: To evaluate the models' alignment quality under specific human preferences.
\end{itemize}

The customized test set contains two parts:
1) real-world sentences derived from online dialogues (2k for each translation direction) that are translated by GPT4 and filtered using a same cleaning strategy described above; and 2) sentences reflecting the same human preferences as the seed preference dataset in English and Spanish (1k for each language) that are translated by human annotators.
The source texts in the customized test set are also from our online services but the source texts are de-duplicate from the training set.

Two public benchmarks are adopted for general translation ability evaluation: 
\begin{itemize}
    \item WMT23 \citep{DBLP:conf/wmt/KocmiABBDFFFGGH23}: WMT is an annually updated benchmark that contains text from various resources (webpages, documents, etc.) and are translated by multiple resources. The latest WMT23 benchmark contains 7 languages: Czech, German, Hebrew, Japanese, Russian, Ukrainian, and Chinese.
    \item Flores200 \citep{DBLP:journals/corr/abs-2207-04672}: The Flores dataset contains parallel data with over 200 languages and the source text are from diverse resources including books, literatures, news, social media, etc.
\end{itemize}

\subsubsection{Metrics}

We use COMET \citep{rei2022comet} and BLEURT \citep{DBLP:conf/acl/SellamDP20} for evaluation and the COMET score is computed by wmt22-comet-da \citep{DBLP:conf/wmt/ReiSAZFGLCM22} with reference.

\begin{figure*}
    \centering
    \begin{minipage}[b]{\textwidth}
        \centering
        \includegraphics[width=0.9\textwidth]{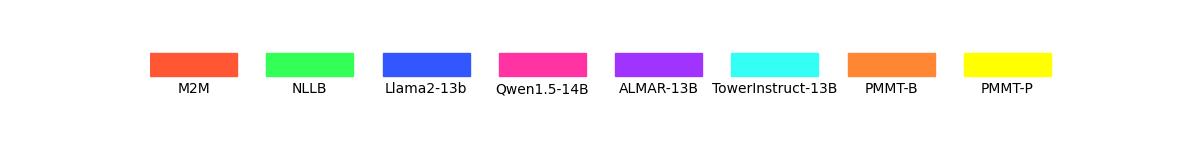}
    \end{minipage}
    \vspace{-1cm} 

    \begin{subfigure}[b]{0.28\textwidth}
        \centering
        \includegraphics[width=\textwidth]{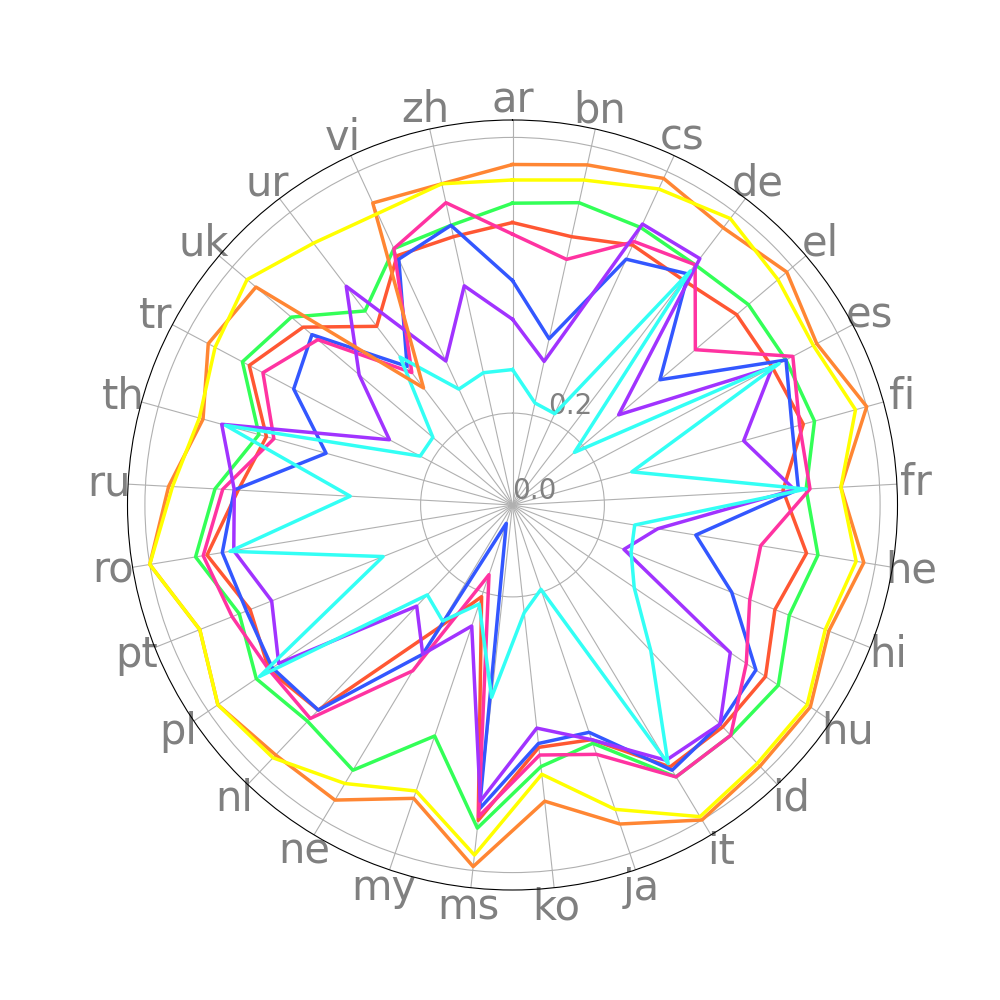}
        \caption{Customized BLEURT (en$\rightarrow$x)}
        \label{fig:customized_bleurt_from_en}
    \end{subfigure}
    \begin{subfigure}[b]{0.28\textwidth}
        \centering
        \includegraphics[width=\textwidth]{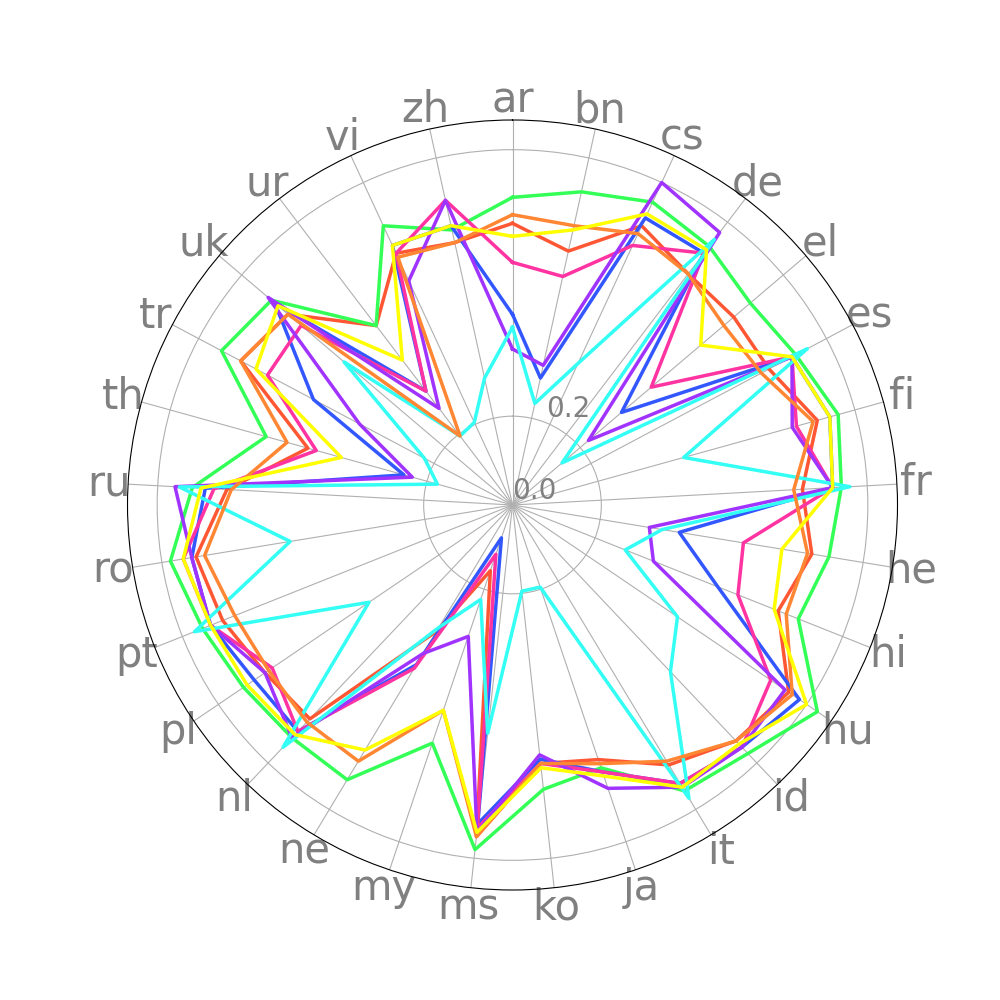}
        \caption{Flores BLEURT (en$\rightarrow$x)}
        \label{fig:flores_bleurt_from_en}
    \end{subfigure}
    \begin{subfigure}[b]{0.28\textwidth}
        \centering
        \includegraphics[width=\textwidth]{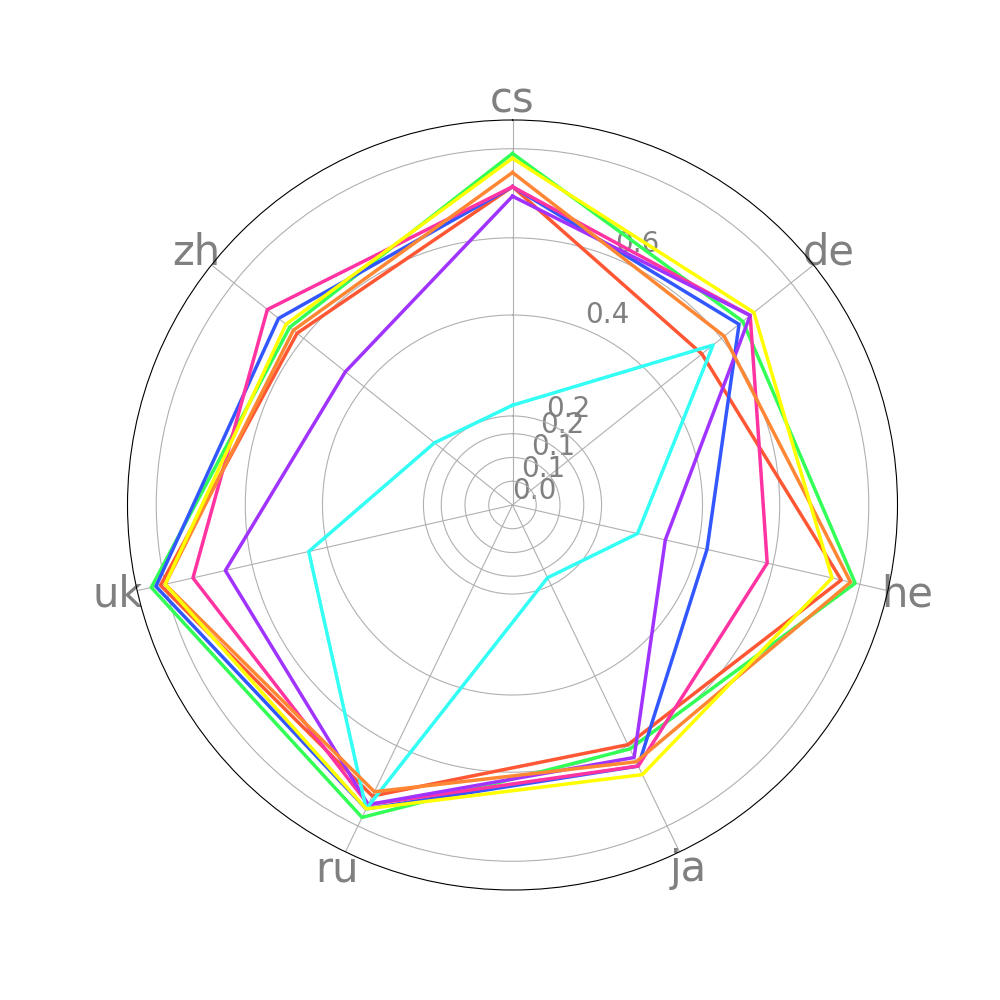}
        \caption{WMT BLEURT (en$\rightarrow$x)}
        \label{fig:wmt_bleurt_from_en}
    \end{subfigure}
    \begin{subfigure}[b]{0.28\textwidth}
        \centering
        \includegraphics[width=\textwidth]{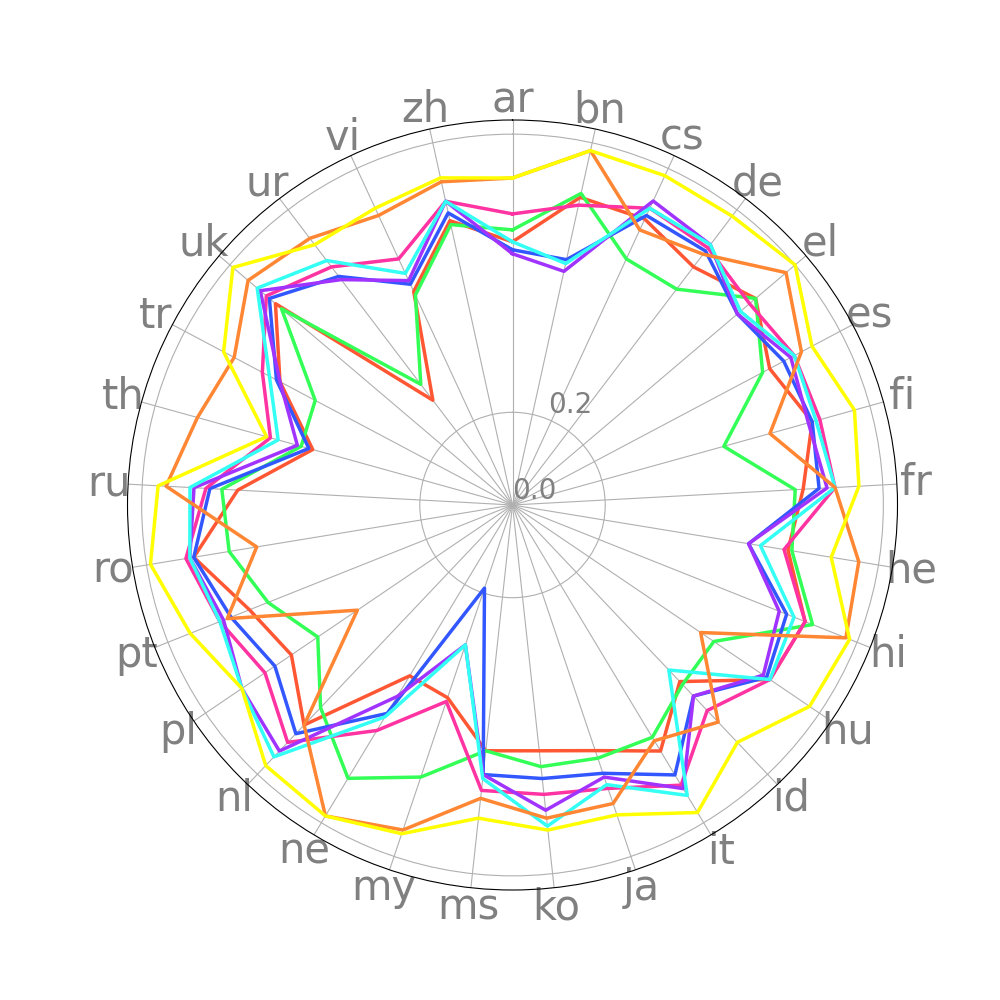}
        \caption{Customized BLEURT (x$\rightarrow$en)}
        \label{fig:ecommerce_bleurt_to_en}
    \end{subfigure}
    \begin{subfigure}[b]{0.28\textwidth}
        \centering
        \includegraphics[width=\textwidth]{E-Commerce_BLEURT-to_EN_radar_chart.png}
        \caption{Flores BLEURT (x$\rightarrow$en)}
        \label{fig:flores_bleurt_to_en}
    \end{subfigure}
    \begin{subfigure}[b]{0.28\textwidth}
        \centering
        \includegraphics[width=\textwidth]{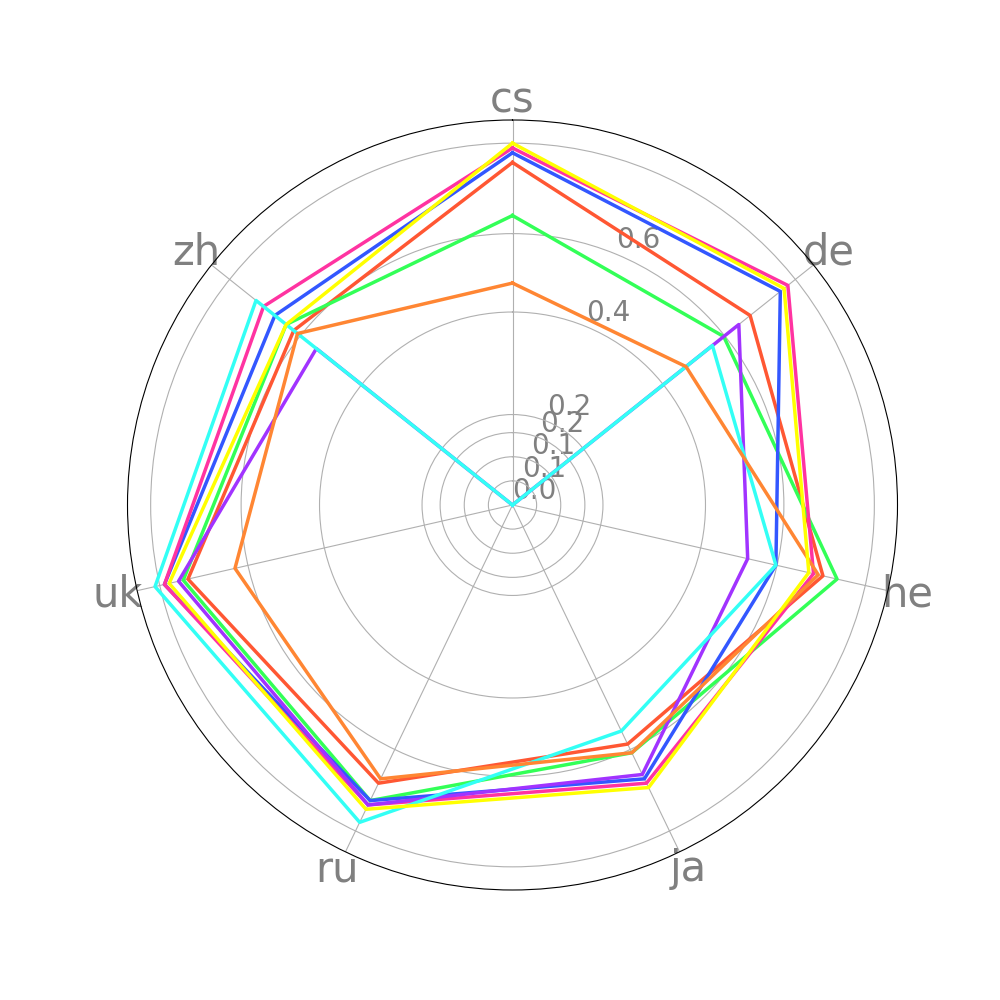}
        \caption{WMT BLEURT (x$\rightarrow$en)}
        \label{fig:wmt_bleurt_to_en}
    \end{subfigure}
    \begin{subfigure}[b]{0.28\textwidth}
        \centering
        \includegraphics[width=\textwidth]{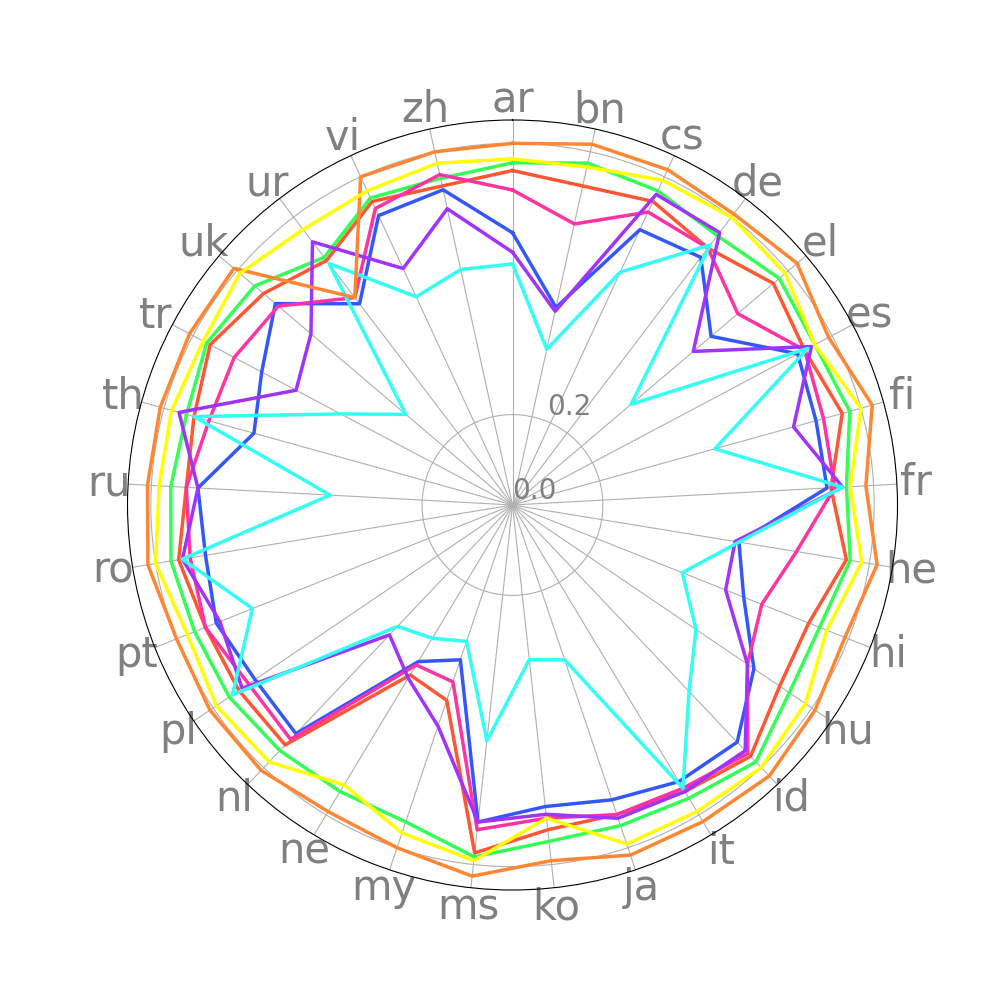}
        \caption{Customized COMET (en$\rightarrow$x)}
        \label{fig:customized_comet_from_en}
    \end{subfigure}
    \begin{subfigure}[b]{0.28\textwidth}
        \centering
        \includegraphics[width=\textwidth]{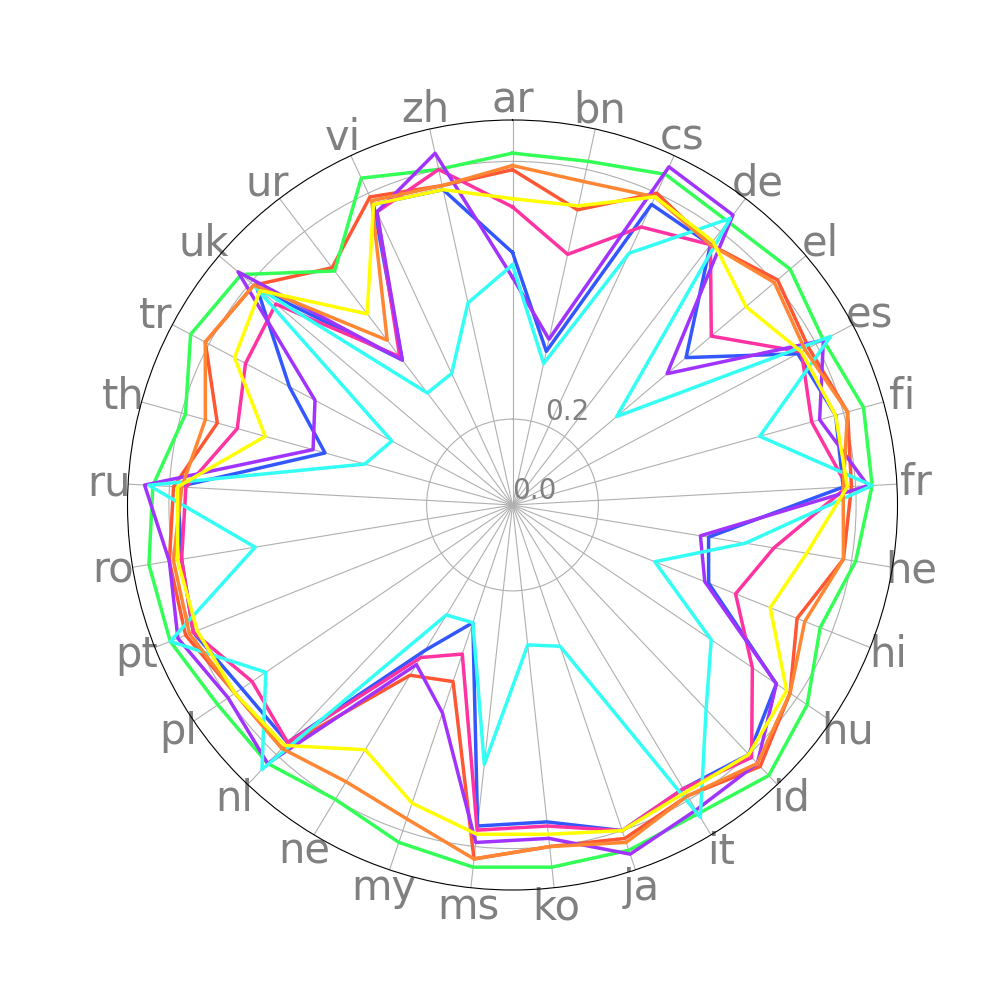}
        \caption{Flores COMET (en$\rightarrow$x)}
        \label{fig:flores_comet_from_en}
    \end{subfigure}
    \begin{subfigure}[b]{0.28\textwidth}
        \centering
        \includegraphics[width=\textwidth]{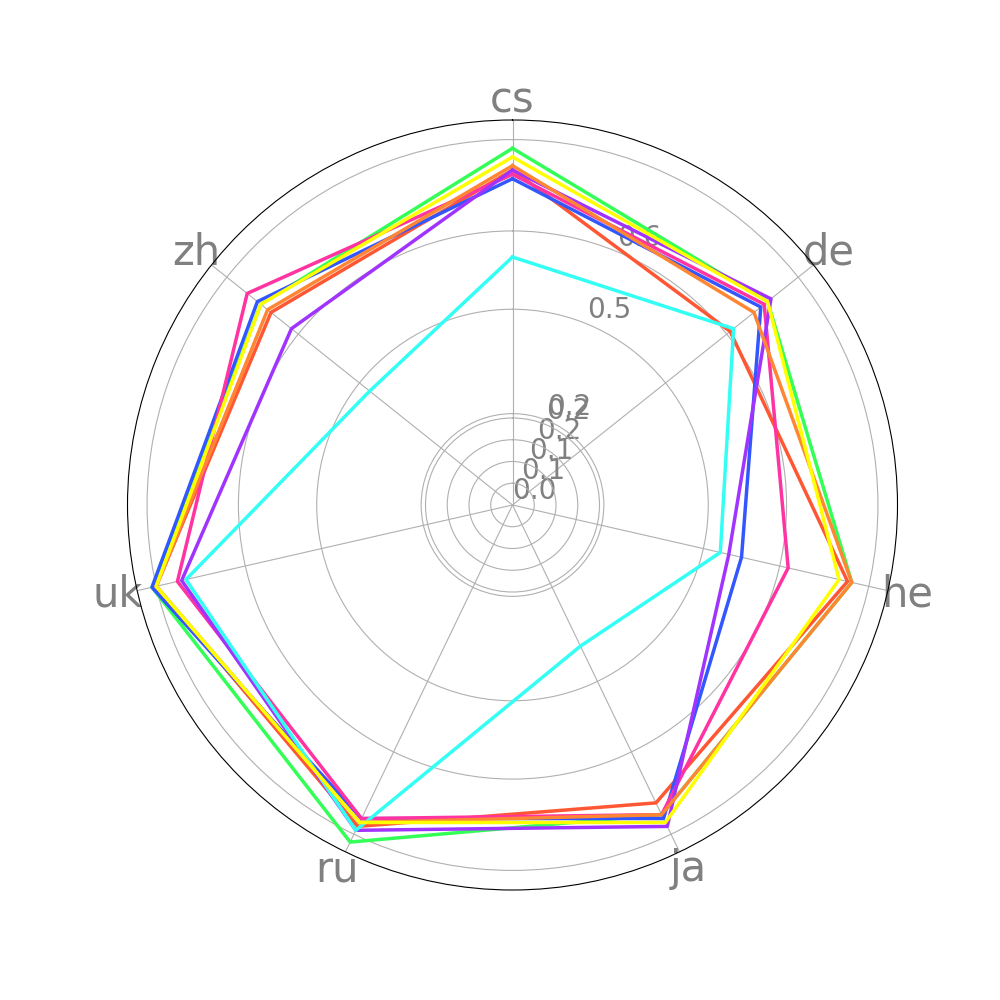}
        \caption{WMT COMET (en$\rightarrow$x)}
        \label{fig:wmt_comet_from_en}
    \end{subfigure}
    \begin{subfigure}[b]{0.28\textwidth}
        \centering
        \includegraphics[width=\textwidth]{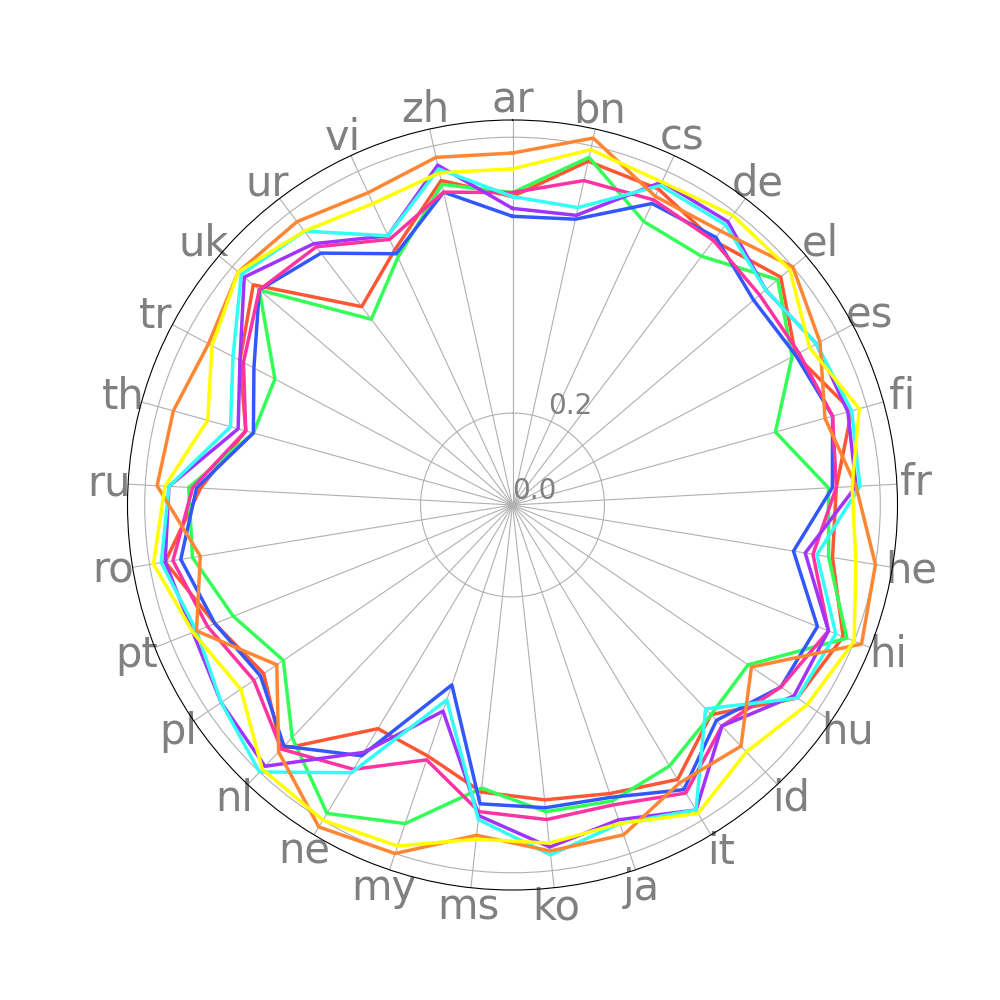}
        \caption{Customized COMET (x$\rightarrow$en)}
        \label{fig:ecommerce_comet_to_en}
    \end{subfigure}
    \begin{subfigure}[b]{0.28\textwidth}
        \centering
        \includegraphics[width=\textwidth]{E-Commerce_COMET-to_EN_radar_chart.png}
        \caption{Flores COMET (x$\rightarrow$en)}
        \label{fig:flores_comet_to_en}
    \end{subfigure}
    \begin{subfigure}[b]{0.28\textwidth}
        \centering
        \includegraphics[width=\textwidth]{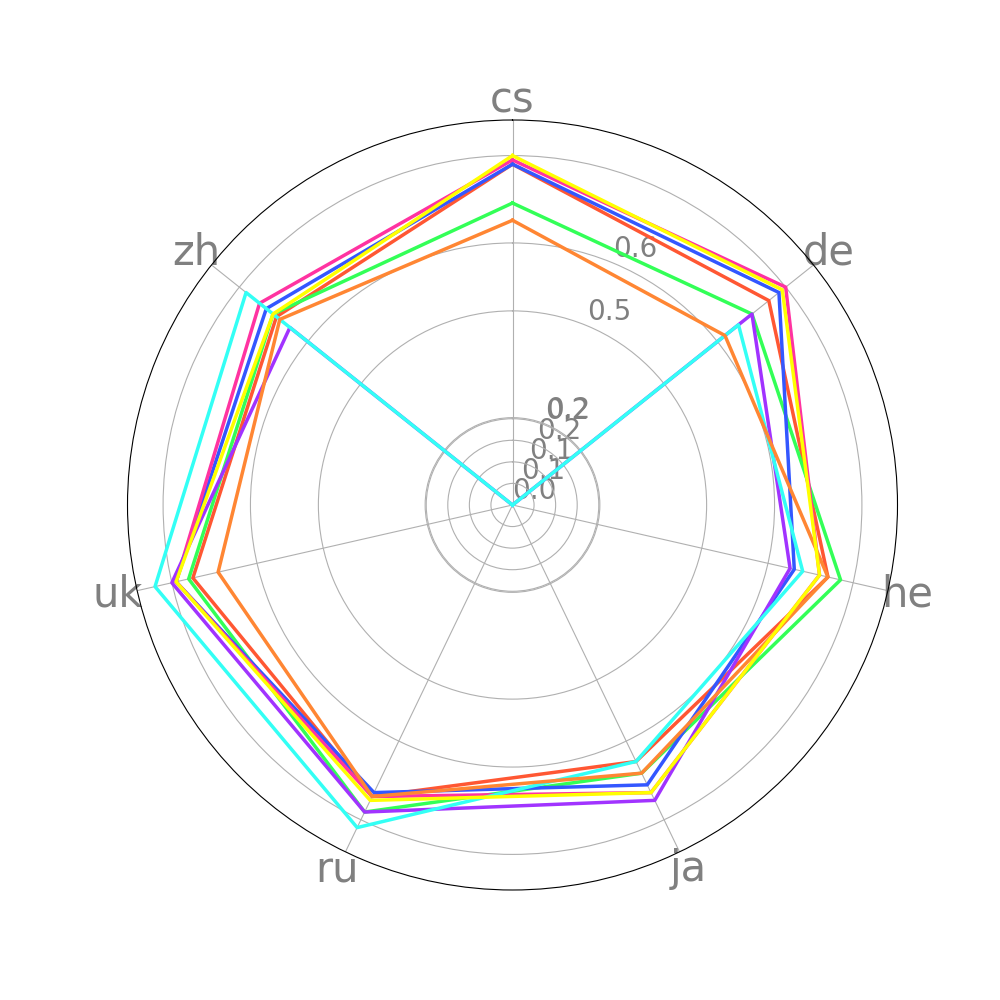}
        \caption{WMT BLEURT (x$\rightarrow$en)}
        \label{fig:wmt_comet_to_en}
    \end{subfigure}
\caption{Detailed comparisons between different translation methods.}
\label{fig:trans_acc}
\end{figure*}

\subsection{Results}

Table.~\ref{tab:trans_acc} illustrates the overall model performances.
Different types of relevant SOTA methods are adopted for a comprehensive comparison: 1) small models like M2M-418M \citep{DBLP:journals/jmlr/FanBSMEGBCWCGBL21} and NLLB-600M \citep{DBLP:journals/corr/abs-2207-04672}; 2) large language models such as LLaMA2-13B \citep{DBLP:journals/corr/abs-2307-09288} and Qwen1.5-14B \citep{DBLP:journals/corr/abs-2309-16609}; and 3) most recent works like ALMA \citep{DBLP:journals/corr/abs-2309-11674} and TowerInstruct \citep{DBLP:journals/corr/abs-2402-17733}. 

For general translation tasks, PMMT-P model surpasses all competitors on WMT23 dataset and maintain a first-class performance on the Flores dataset.
Although the PMMT-B model slightly left behind comparing to PMMT-P, its translation accuracy is still competitive to many models that are much larger.
Please note that both PMMT models are not trained on Flores nor WMT. Meanwhile, the text sources of the seed data are dramatically different from these two benchmarks.
As a result, the evaluation scores under the public datasets proves the generalization ability of the proposed method.

\begin{table*}
    \centering
    \begin{tabular}{r|c|c|c|c|c|c|c}
        \hline
        \multirow{2}{*}{Models} & \multicolumn{2}{c|}{pref\_data} & \multicolumn{2}{c|}{common\_data} & \multicolumn{2}{c|}{flores} & \multirow{2}{*}{Win Rate} \\
        \cline{2-7}
        ~ & BLEURT & COMET & BLEURT & COMET & BLEURT & COMET & ~ \\
        \hline
        SFT Model 1 & 0.86 & 0.90 & \textbf{0.87} & \textbf{0.92} & \textbf{0.75}  & \textbf{0.86} & 4/6 \\
        SFT Model 2 & 0.86 & 0.90 & 0.86 & 0.91 & 0.74  & 0.85 & 0/6 \\
        SFT Model 3 & 0.85 & 0.89 & \textbf{0.87} & \textbf{0.92} & \textbf{0.75}  & \textbf{0.86} & 4/6 \\
        \hline
        DPO Model & \textbf{0.88} & \textbf{0.91} & \textbf{0.87} & \textbf{0.92} & \textbf{0.75}  & \textbf{0.86} & \textbf{6/6} \\
        PPO Model & 0.87 & 0.90 & \textbf{0.87} & \textbf{0.92} & 0.74  & 0.85 & 2/6 \\
        \hline
    \end{tabular}
    \caption{Comparison between different training strategies. Best results are highlighted in \textbf{bold}.}
    \label{tab:dpo-ppo}
\end{table*}

On the customized test set, performances of both PMMT models take the lead by a large margin. This result indicates that both PMMT models have learnt the customized human preferences and can generate satisfactory translations.

By comparing results across both public datasets and the customized dataset, we can draw the conclusion that the proposed method can align to human preferences while maintaining the general translation accuracy.
Meanwhile, the PMMT-P model outperforms the PMMT-B model in almost all benchmarks, which validates that using PMMT-P to generate the candidate translations and training PMMT-B models on the generated data can improve the performance of PMMT-B models along automatic updates.

Detailed results of each translation direction of all languages are shown in Figure.~\ref{fig:trans_acc}, which generally match the above conclusions.
Additionally, by comparing the performances for each language, we find that the relative performance of PMMT-B models mainly follows the ability of its base model under public datasets, while improving performance on the customized dataset.

\subsection{Ablation Study}

\subsubsection{Training strategy}

Ablation studies are conducted to test the effectiveness of our training strategy by comparing SFT with DPO \citep{DBLP:conf/nips/RafailovSMMEF23} and PPO \citep{DBLP:journals/corr/SchulmanWDRK17}.
We chose Spanish-English translation task as an example and experiments are performed on the Flores dataset and our customized test set since WMT23 does not have Spanish.
To evaluate the influence of general translation and translation with human preferences, we split the customized dataset into two parts: 
\begin{itemize}
    \item common samples (trainset, 256k) and common\_data (test set, 30k): samples randomly selected from the seed dataset;
    \item preference samples (trainset, 20k) and pref\_data (test set, 2k): samples randomly selected from the seed preference dataset.
\end{itemize}

Performances of different training strategies are shown in Table.~\ref{tab:dpo-ppo}. Training details of each model is listed below:
\begin{itemize}
    \item SFT Model 1 (PMMT-P strategy): Simultaneously trained on common samples and preference samples for 5 epochs.
    \item SFT Model 2: Successively trained on common samples for 5 epochs and then fine-tuned on preference samples for 1 epoch.
    \item SFT Model 3: Trained on common samples for 5 epochs.
    \item DPO Model: Trained on preference samples for 1 epoch after SFT using DPO strategy.
    \item PPO Model: Trained on preference samples for 1 epoch after SFT using PPO strategy.
\end{itemize}

In general, although DPO performs best in all test sets, the performance gap between strategies is very small.
Considering that the training cost of SFT is much less than DPO and PPO, we trained the PMMT-P model with the SFT strategies.
Among them, the strategy used to train SFT Model 1 is chosen since it provides the best performance.

\subsubsection{Preference Data Scale and Model Size}

\begin{figure}
    \includegraphics[width=\linewidth]{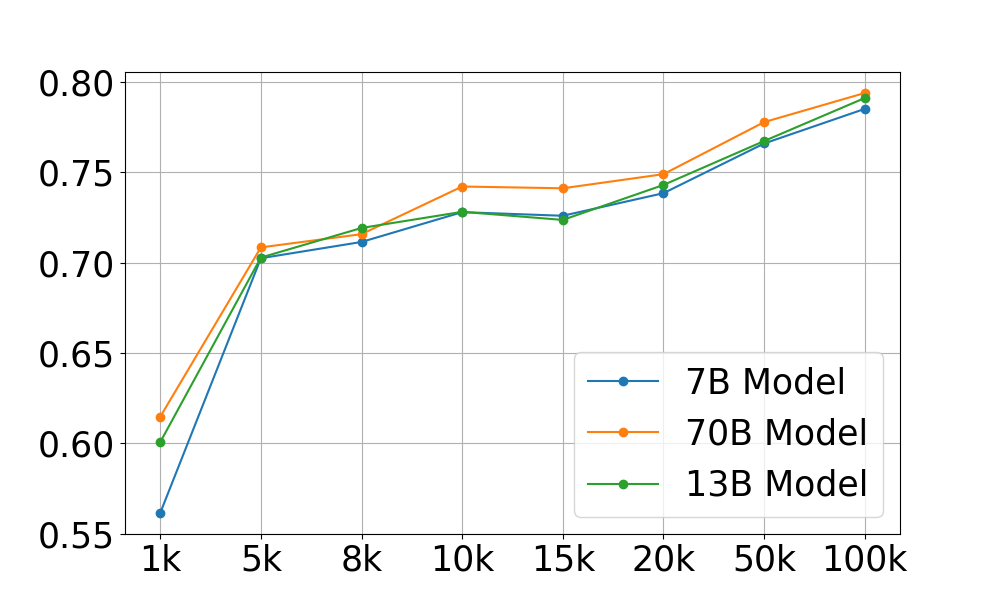}
    \caption{Accuracy of the PMMT-J model trained with different data and model scales.}
    \label{fig:jm_data_scale}
\end{figure}

An ablation study was conducted to determine the optimal data scale for PMMT-J training.
Specifically, we trained the PMMT-J model with 7B, 13B, and 70B parameters on datasets with scales ranging from 1k to 100k entries in English and Spanish.
As shown in Figure.~\ref{fig:jm_data_scale}, different model scales follow a similar trend while there is an obvious plateau between data scales of 5k and 20k, where accuracy improvements are minimal despite the increasing data scale.
Consequently, a data scale of 10k was selected to train the PMMT-J model with 13B parameters to balance training cost and performance.

\section{Conclusion}

In this paper, a novel method is proposed to align customized human preferences in multilingual translation, and knowledge from LLMs is distilled to smaller models automatic for more effective online serving. 
First, a small-scale seed dataset is produced to train the translation LLM for generating candidate translations with various styles. 
Then, an RM is trained on another seed preference dataset and is used to select translations that match human preferences from the candidates generated by the LLM. 
Finally, large-scale multilingual corpora are generated and used to train the smaller translation analyzing models, which can be updated effectively with the proposed pipeline.
Experimental results prove that our method can not only align with human preferences but also maintain general translation accuracy.

\bibliography{custom}

\section{Reproducibility Checklist}
Unless specified otherwise, please answer “yes” to each question if the relevant information is described either in the paper itself or in a technical appendix with an explicit reference from the main paper. If you wish to explain an answer further, please do so in a section titled “Reproducibility Checklist” at the end of the technical appendix.

This paper:

Includes a conceptual outline and/or pseudocode description of AI methods introduced (yes)

Clearly delineates statements that are opinions, hypothesis, and speculation from objective facts and results (yes)

Provides well marked pedagogical references for less-familiare readers to gain background necessary to replicate the paper (yes)

Does this paper make theoretical contributions? (yes)

If yes, please complete the list below.

All assumptions and restrictions are stated clearly and formally. (yes)

All novel claims are stated formally (e.g., in theorem statements). (yes)

Proofs of all novel claims are included. (yes)

Proof sketches or intuitions are given for complex and/or novel results. (yes)

Appropriate citations to theoretical tools used are given. (yes)

All theoretical claims are demonstrated empirically to hold. (yes)

All experimental code used to eliminate or disprove claims is included. (yes)

Does this paper rely on one or more datasets? (yes)

If yes, please complete the list below.

A motivation is given for why the experiments are conducted on the selected datasets (yes)

All novel datasets introduced in this paper are included in a data appendix. (partial: We public the customized test set)

All novel datasets introduced in this paper will be made publicly available upon publication of the paper with a license that allows free usage for research purposes. (yes)

All datasets drawn from the existing literature (potentially including authors’ own previously published work) are accompanied by appropriate citations. (yes)

All datasets drawn from the existing literature (potentially including authors’ own previously published work) are publicly available. (yes)

All datasets that are not publicly available are described in detail, with explanation why publicly available alternatives are not scientifically satisficing. (yes)

Does this paper include computational experiments? (yes)

If yes, please complete the list below.

Any code required for pre-processing data is included in the appendix. (no: This is trade secret. But we have make it clear enough for reproducing the method including methods and tools that we used.).

All source code required for conducting and analyzing the experiments is included in a code appendix. (no. The inference code has been described in the paper, which is publicly available. The metrics used are stadard critera and computation method is also publicly available. Checkpoints of our models are trade secret.)

All source code required for conducting and analyzing the experiments will be made publicly available upon publication of the paper with a license that allows free usage for research purposes. (partial)

All source code implementing new methods have comments detailing the implementation, with references to the paper where each step comes from (yes)

If an algorithm depends on randomness, then the method used for setting seeds is described in a way sufficient to allow replication of results. (yes)

This paper specifies the computing infrastructure used for running experiments (hardware and software), including GPU/CPU models; amount of memory; operating system; names and versions of relevant software libraries and frameworks. (partial. We have written all computing infrastructure that is necessary information to reproduce the method.)

This paper formally describes evaluation metrics used and explains the motivation for choosing these metrics. (yes)

This paper states the number of algorithm runs used to compute each reported result. (yes)

Analysis of experiments goes beyond single-dimensional summaries of performance (e.g., average; median) to include measures of variation, confidence, or other distributional information. (yes)

The significance of any improvement or decrease in performance is judged using appropriate statistical tests (e.g., Wilcoxon signed-rank). (yes)

This paper lists all final (hyper-)parameters used for each model/algorithm in the paper’s experiments. (partial. We have listed all necessary hyper-parameters.)

This paper states the number and range of values tried per (hyper-) parameter during development of the paper, along with the criterion used for selecting the final parameter setting. (yes)

\end{document}